\documentclass[conference]{IEEEtran}
\IEEEoverridecommandlockouts
\usepackage{cite}
\usepackage{amsmath,amssymb,amsfonts}
\usepackage{algorithmic}
\usepackage{graphicx}
\usepackage{textcomp}
\usepackage{xcolor}

\usepackage[utf8]{inputenc}
\usepackage{algorithm}
\usepackage{algorithmic}
\usepackage{multirow}
\usepackage{graphicx}

\def\BibTeX{{\rm B\kern-.05em{\sc i\kern-.025em b}\kern-.08em
    T\kern-.1667em\lower.7ex\hbox{E}\kern-.125emX}}
\begin{document}

\title{Federated Split BERT for Heterogeneous Text Classification}


\author{\IEEEauthorblockN{Zhengyang Li$^{\dagger}$, Shijing Si$^{\dagger}$, Jianzong Wang$^{*}$ and Jing Xiao}
   \IEEEauthorblockA{\textit{Ping An Technology (Shenzhen) Co., Ltd., Shenzhen, China}
  \\ Email:  \{lizhengyang927, sishijing204, wangjianzong347, xiaojing661\}@pingan.com.cn}
\thanks{$^{\dagger}$ Equal contribution.}
 \thanks{* Corresponding author is Jianzong Wang.}
 }

\maketitle

\begin{abstract}
Pre-trained BERT models have achieved impressive performance in many natural language processing (NLP) tasks. However, in many real-world situations, textual data are usually decentralized over many clients and unable to be uploaded to a central server due to privacy protection and regulations. Federated learning (FL) enables multiple clients collaboratively to train a global model while keeping the local data privacy. A few researches have investigated BERT in federated learning setting, but the problem of performance loss caused by heterogeneous (e.g., non-IID) data over clients remain under-explored. 
To address this issue, we propose a framework, FedSplitBERT, which handles heterogeneous data and decreases the communication cost by splitting the BERT encoder layers into local part and global part. The local part parameters are trained by the local client only while the global part parameters are trained by aggregating gradients of multiple clients. Due to the sheer size of BERT, we explore a quantization method to further reduce the communication cost with minimal performance loss. Our framework is ready-to-use and compatible to many existing federated learning algorithms, including FedAvg, FedProx and FedAdam. Our experiments verify the effectiveness of the proposed framework, which outperforms baseline methods by a significant margin, while FedSplitBERT with quantization can reduce the communication cost by $11.9\times$.


\end{abstract}


\section{Introduction}

Lately language model pre-training has shown to be highly effective in learning universal language representations from large-scale unlabeled data. Pre-trained models such as ELMo \cite{peters2018deep}, GPT \cite{brown2020language} and BERT \cite{devlin2019bert} have
achieved great success in many NLP tasks, such as sentiment classification \cite{wang2016attention}, natural language inference \cite{williams2018broad}, and question answering \cite{lai2017race}.
In many real-world applications, textual data such as clinical records are decentralized and stored locally on client devices \cite{xu2021federated}, such as phones and personal computers. Due to the recent stringent regulation on private data protection \cite{voigt2017eu}, these data cannot be directly uploaded to a central server. Federated learning \cite{mcmahan2017communication,kairouz2021advances} allows multiple clients to train a global deep neural network model collaboratively in a distributed environment without moving data to a centralized storage.


There have been some attempts on fine-tuning BERT in federated learning setting. For example, \cite{huang2020texthide} proposed an encryption method for clients to protect data privacy while keeping the performance of fine-tuned models comparable to centralized fine-tuning.
\cite{liu2020federated} confirms that it is possible to both pre-train and fine-tune BERT models in a federated manner using clinical texts from different silos without moving the data. \cite{hilmkil2021scaling} provides an overview of the applicability of the federated learning to Transformer-based language models. \cite{wang2021modeling} firstly applied federated learning to Transformer-based Neural Machine Translation to avoid sharing customers' chat recording with the server.

However, there are still a few under-explored challenges for fine-tuning BERT over decentralized data: firstly, 
the data distribution on clients may vary substantially, which might cause severe performance degradation of the fine-tuned model \cite{yang2020heterogeneity}; secondly, due to the huge number of parameters in BERT, the communication cost between clients and the central server can be very high. 

To address these concerns, we propose a new framework, FedSplitBERT, referred to as \textbf{Fed}erated \textbf{Split} \textbf{BERT} for heterogeneous data. Specifically, we split the whole BERT encoders (12 layers for BERT Base or 24 layers for BERT Large) into two parts at one specific layer, called \textit{critical layer}. BERT encoder layers above this critical layer (excluding itself) are locally learned with data on this client only, rather than jointly trained over all clients. Encoder layers below the critical layer (including itself) are jointly trained over all clients.
The intuition behind our method is that encoder layers above the critical layer are close to the softmax output, so their parameters can be tuned to adapt the local data distribution. The layers below the \textit{critical layer} are used as a shared feature extractor over all clients (global part). Instead of being a provider of data during the FL, every client finally trains a personalised model consisting of global part and loacl part which satisfied its specific data distribution. Moreover, because only a part of the parameters need to be transmit between central server and end-point clients, the communication cost during model training will drop sharply.

\vspace{2.5pt}
Our contributions are summarized as follows:
\begin{itemize}
   \item We propose a new federated learning framework, FedSplitBERT, which specializes in fine-tuning BERT on heterogeneous data over multiple clients.
   \item We further investigate quantization method to reduce the communication cost.
   \item We conduct extensive experiments to show the effectiveness of our methods and prove that the split architecture of the Transformer-based model has the great ability to tackle the heterogeneity problems.
\end{itemize}

\section{Related Works}

\subsection{Federated Learning}
The general FL setup involves two types of updates, the server and local, the local updates are associated with minimizing some local loss function, while the global updates aggregate local weights and sychronize the updated model at each communication round. 

Federated averaging (FedAvg) \cite{mcmahan2017communication} and federated proximal (FedProx) \cite{li2020federated} are two general-purpose algorithms in federated learning. FedAvg is an iterative method that each client $i$ optimizes a local surrogate of global objective function, and merges the weights $W_i$ at each communication round $t$, with the same learning rate $\eta$ on each client. 
To address the heterogeneity issue, FedProx adds a regularization term to force the weights $W_i$ on client $i$ to be closer to global model $W$ during training. FedProx reaches more stable and better performance than FedAvg on images classification tasks. Recent advanced methods like FedOpt \cite{reddi2021adaptive} and FedDyn \cite{acar2021federated} follow similar pipeline while improve model aggregation and reduce communication cost. FedSmart \cite{FedSmart111} optimizes the performance of the client model on its local data by updating weights based on the accuracy of a local validation set. These model aggregation ideas are complementary to our work and they can be integrated in the global update step in our proposed framework.

Another line of FL works aim to decrease communication cost by splitting the whole model architecture into two parts: the local model and global model \cite{liang2020think, yang2020heterogeneous, collins2021exploiting}. The main idea lies in that each client keeps a local model which can adapt to local data to mitigate the effect of data heterogeneity as well as reduce the amount of parameters need to be transmitted.
For instance, \cite{dinh2020personalized} formulates a new bi-level optimization problem designed for personalized FL (pFedMe) by using the Moreau envelope as a regularized loss function. This work can produce personalized models for individual clients, but its communication cost might be too high for the BERT model. Our FedSplitBERT improves communication efficiency by learning part of BERT layers locally and quantization. Local global FedAvg (LG-FedAvg) \cite{liang2020think} and the heterogeneous Data Adaptive Federated Learning (HDAFL) algorithm \cite{yang2020heterogeneous} utilizes similar strategies as us: (1) locally train client specific neural network layers (local models) and (2) aggregate generic (global) model parameters shared by all clients. 
However, these two papers mainly focus on architectures like convolution neural networks (CNN) and multi-layered perceptrons (MLPs), rather than the Transformer-based pre-trained BERT models. Our paper mainly studies how to fine-tune BERT in a federated setting over heterogeneous data on clients.

\subsection{BERT and Quantization}
Since its inception, BERT and its variants have been the base models for many NLP applications \cite{devlin2019bert,si2020students,wang2020integrating}. There are a few research on applying BERT in the federated setting, such as \cite{liu2020federated} on federated pre-training and fine-tuning BERT over decentralized clinical notes.
However, how to fine-tune BERT on heterogeneous textual data remains under-explored. In this work, we will fill this gap by proposing FedSplitBERT. Due to the large number of parameters in BERT models, the communication efficiency is a significant challenge for FL.


Quantization is a commonly used approach to reduce model size while maximally keeping its performance \cite{shen2020q,liu2021hardware}. It reduces the number of bits used to represent a number in the model. For instance, \cite{shen2020q} proposes a Hessian based ultra low precision quantization method for BERT. \cite{kim2021bert} study the integer and binary quantization methods for the BERT model during the inference phase, respectively. \cite{zafrir2019q8bert} achieves the impressive compression ratio by using quantization-aware training during fine-tuning process and quantizing all the Embedding layers and Fully Connected layers of BERT to 8-bit during inference. \cite{jin2021kdlsq} leverages the learned step size quantization for Transformer-based model, i.e. BERT, and further uses the knowledge distillation technique to get a compressed "student" BERT model. \cite{reisizadeh2020fedpaq} adapts a random Low-precision quantizer for gradients to federated learning manner. These methods can be used in our FedSplitBERT and could potentially reduce communication cost in the federated BERT fine-tuning.

\section{Methodology}
This section covers the details of our FedSplitBERT framework and how we tackle the challenges of heterogeneous data and communication bottleneck in the federated setting.

\subsection{FedSplitBERT}
\begin{figure}[htb]
    \centering
    \includegraphics[width=0.5\textwidth]{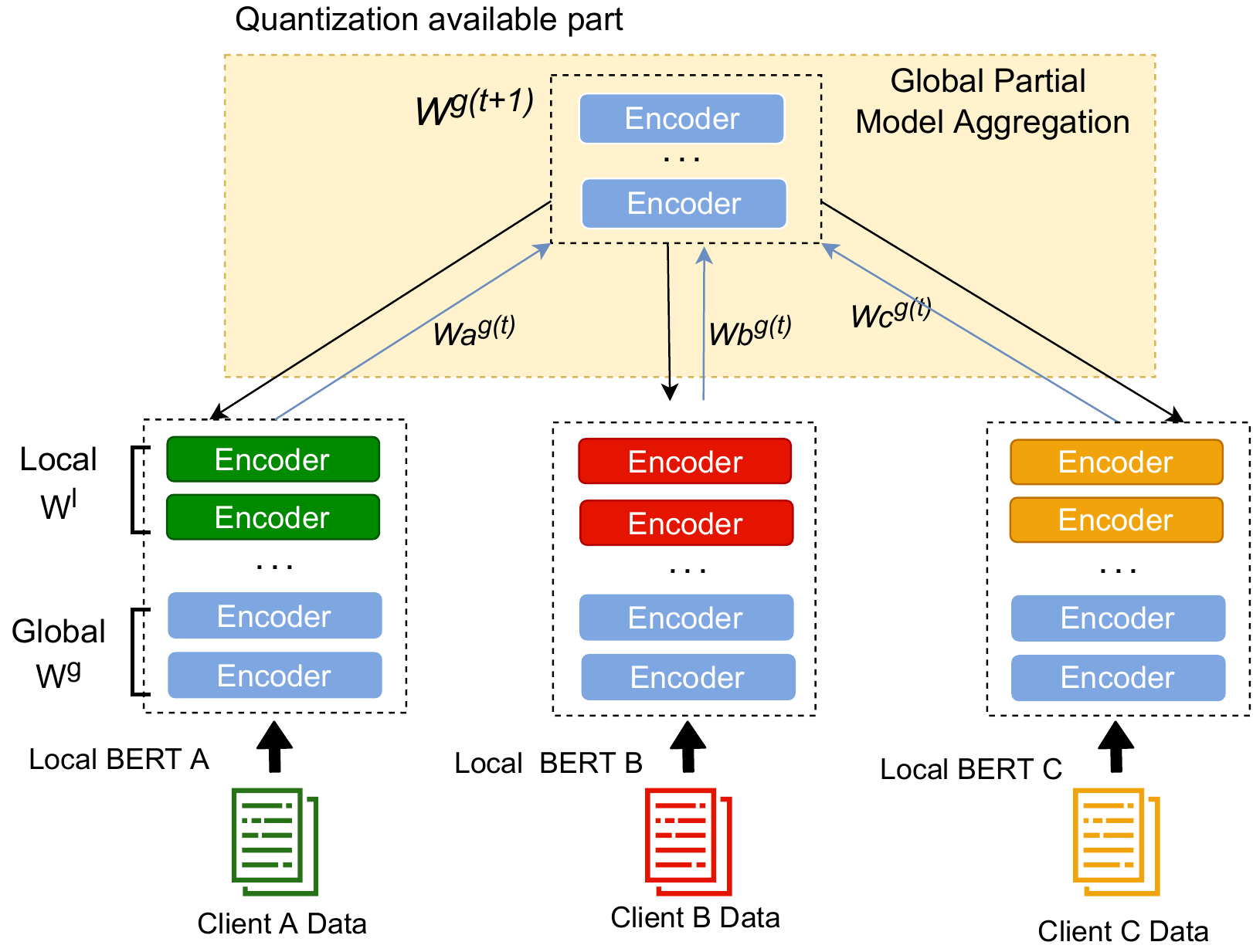}
    \caption{Pictorial view of the proposed FedSplitBERT in federated learning for the 12-layer BERT model. All clients share a set of encoder layers from Layer 1 to Layer $c$ (colored blue) and have distinct local layers (from Layer $c+1$ onward) that can potentially adapt to individual clients. The global layers are shared with the server while the local layers are kept private by each client. 
}
\label{fig:fedbert}
\end{figure}

\renewcommand{\algorithmicrequire}{\textbf{Server executes}}  
\renewcommand{\algorithmicensure}{\textbf{ClientUpdate}$(i,\mathbf{W}^{g(t)})$}

\begin{algorithm}[htb]
\setlength{\baselineskip}{13pt}
\caption{FedSplitBERT with quantization. \\ $\eta$ is the learning rate, the total sample size $N=\sum_{i=1}^{K}N_i$, $N_i$ denotes the sample size on client $i$, $\text{qat}(\cdot)$ is the quantization function}
\label{alg:fed.bert}
\begin{algorithmic}[1] 
\REQUIRE:
\STATE Initialize global model $\mathbf{W}^{g(0)}$ as a 16-bit tensor and local models on $K$ clients $\mathbf{W}^{l(0)}_{i}$ as 32-bit tensors
\FOR{iteration $t = 0, 1, \ldots, T$}
\STATE $S\longleftarrow$(randomly select a number of clients from $K$)
\FOR{$i\in S$ in parallel}
\STATE $\mathbf{W}^{g(t+1)}_{i}\leftarrow\text{ClientUpdate}(i, \mathbf{W}^{g(t)})$
\ENDFOR
\STATE $\mathbf{W}^{g(t+1)}\longleftarrow\sum_{i=1}^{K}\frac{N_{i}}{N}\text{qat}(\mathbf{W}_{i}^{g(t+1)})$
\ENDFOR
\end{algorithmic}
\begin{algorithmic}[1]
\ENSURE 
\STATE Cast the 16-bit global tensor $\mathbf{W}^{g(t)}$ to 32-bit on client $i$ as $\mathbf{W}^{g(t)}_{i}$
\STATE Combine $\mathbf{W}^{g(t)}_{i},\mathbf{W}^{l(t)}_{i}$ into $W^{(t)}_i$
\STATE Split local data into a set of mini-batches $\mathcal{B}$
\FOR{each mini-batch $b$ $\in\mathcal{B}$}
\STATE Optimize $\mathcal{L}(x_b,y_b;W^t_i)$ with local batch to compute $\nabla W^{g(t+1)}_{i}, \nabla W^{l(t+1)}_{i}$
\STATE $\mathbf{W}^{g(t+1)}_{i}\longleftarrow\mathbf{W}^{g(t)}_{i}-{\eta}\nabla W^{g(t+1)}_{i}$
\STATE $\mathbf{W}^{l(t+1)}_{i}\longleftarrow\mathbf{W}^{l(t)}_{i}-{\eta}\nabla W^{l(t+1)}_{i}$
\ENDFOR
\STATE Quantize $\mathbf{W}^{g(t+1)}_{i}$ to 16-bit 
\STATE return $\text{qat}(\mathbf{W}^{g(t+1)}_{i})$
\end{algorithmic}
\end{algorithm}

Fig. \ref{fig:fedbert} depicts how our FedSplitBERT proceeds: for each client, it splits the whole BERT encoders into local and global encoder layers.
All clients share a set of encoder layers from Layer 1 to Layer $c$ (including itself), called the global model, and each client has distinct local layers (from Layer $c+1$ onward) that can potentially adapt to heterogeneous (non-identically independent distributed) client data. Specifically, for the $i$-th client, we denote its parameters as $\mathbf{W}_{i} = [\mathbf{W}^{g}, \mathbf{W}^{l}_i]$, where $\mathbf{W}^{g}$ is the global parameters in the shared encoder layers, and $\mathbf{W}^{l}_i$ is the local model parameters specific to the $i$-th client.

During fine-tuning of the $t$-iteration, server randomly select $s$ clients ($s\in\{1,K\}$) among all active participants, $K$ is the total number of clients. the selected $i$-th client feeds each batch of data to the global layers first to extract features, then pass their features to the local layers to compute the cross-entropy loss. The difference between local and global layers is that after back propagation, client $i$ only uploads its calculated global weights $\mathbf{W}^{g(t+1)}_{i}$ to the central server, then the server performs aggregation and synchronizes the new weights of global encoders $W^{g(t+1)}$ to clients.
Theoretically, our goal is to optimize the function: 
\begin{equation}
\mathop{\min}\limits_{\mathbf{W}^{g}, \mathbf{W}^{l}_{1}, \ldots, \mathbf{W}^{l}_{K}}\mathcal{L}(\cdot)=\frac{1}{K}\sum_{i=1}^{K}\mathcal{L}_i(x_i,y_i; \mathbf{W}_i)
\end{equation}
$\mathcal{L}_i(\cdot)$ is the loss of the prediction on $(x_i,y_i)$, where $(x_i,y_i)$ is the data on the $i$-th client.
The FedSplitBERT is presented in Algorithm \ref{alg:fed.bert}, and here we use FedAvg to aggregate client weights. Actually, other aggregation methods like FedOpt could also be used.

In FedSplitBERT, global encoders from Layer 1 to Layer $c$ act as a common feature extractor to capture surface features\cite{jawahar2019does}, while the local encoders (from Layer $c$ onward) aim to extract high-level features that reflect the individual characteristics. We call Layer $c$ the \emph{critical layer}, and in federated BERT fine-tuning, we treat $c$ as a hyper-parameter and it can balance the overfitting and communication cost of the system. For a fixed BERT architecture, when $c$ is small, the BERT model is mostly an individual local model for each client, the local model will have large number of parameters (and tends to overfitting the data), but the communication cost will be low. An extreme case is $c=0$, where every client has its independent BERT model trained on its own isolated data and there is no communication between clients and the server. On the contrary, when $c$ is large and close to the top layers, the local model on each client is small and has small number of parameters (less likely to overfitting the local data), but the communication cost is high. An extreme case of this way is when $c=12$ for the standard BERT model (12 layers in total), all clients share the same BERT model, equivalent to FedAvg algorithm. The problem of FedAvg is that its performance degrades substantially for heterogeneous clients.
In our experiments, we treat $c$ as a hyper-parameter and it can be tuned to choose the best model for each dataset. 

\subsection{Quantizing Global Weights}

FedSplitBERT only requires clients to upload and download $\mathbf{W}^{g}$ (i.e. the weights of the global model) in each communication round, so the communication cost is significantly reduced. Considering the huge number of parameters in BERT, there might still be millions of parameters to upload or download at each communication round and the network costs are still prohibitive for portable edge devices.

\begin{figure}[htb]
    \centering
    \includegraphics[width=0.5\textwidth]{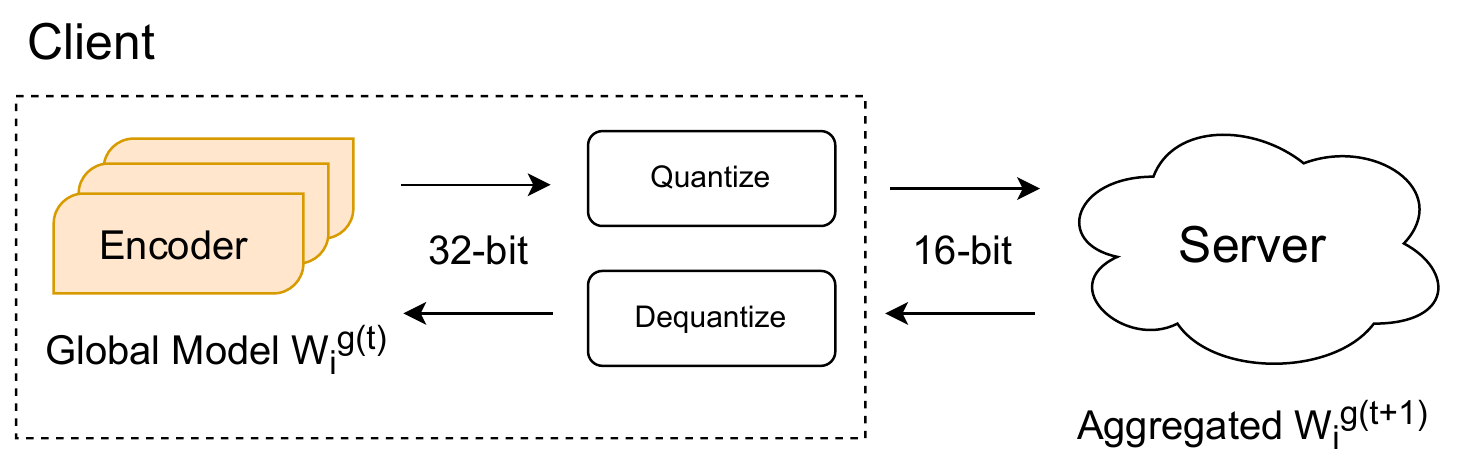}
    \caption{Quantization Scheme of FedSplitBERT}
\label{fig:quantize_fig}
\end{figure}

To further improve communication efficiency, we employ quantization on all the parameters to upload and download (See the quantization process in Fig. \ref{fig:quantize_fig}). For $s$ selected clients in each iteration, it proceeds in the following steps: (1) after the back propagation, quantize the global weights $\mathbf{W}_{i}^{g}$ from 32-bit to 16-bit on client $i$; (2) upload the quantized $\mathbf{W}_{i}^{g}$ to the server for aggregation; (3) the server aggregate quantized weights in 16-bit according to a certain rule (it could be FedAvg, FedProx or FedOpt, etc); (4) selected clients download the updated $\mathbf{W}^{g}$ and inversely quantize $\mathbf{W}^{g}$ from 16-bit to 32-bit.
The whole FedSplitBERT algorithm with quantization is illustrated as Algorithm \ref{alg:fed.bert}.

\section{Experiments}
Here we conduct experiments to verify the effectiveness of FedSplitBERT on text classification over extensive public datasets. We aim to demonstrate that our methods can (1) fine-tune BERT over multiple clients with heterogeneous data and produce better performance than baseline methods like FedAvg, FedProx and FedAdam, and (2) reduce the communication cost during fine-tuning in terms of uploading and downloading size. 

\subsection{Baseline Methods \& Datasets}

The baseline methods included in this paper are: FedAvg, FedProx and FedAdam \cite{reddi2021adaptive}. 
We evaluate all methods on General Language Understanding Evaluation (GLUE) benchmark \cite{wang2018glue}, a collection of 9 sentence-level language understanding tasks:
\begin{itemize}
    \item Two sentence-level classification tasks including Corpus of Linguistic Acceptability
(CoLA) \cite{warstadt2019neural}, and Stanford
Sentiment Treebank (SST-2) \cite{socher2013recursive}.
\item Three sentence-pair similarity tasks including Microsoft Research Paraphrase Corpus (MRPC) \cite{dolan2005automatically}, Semantic Textual Similarity Benchmark (STSB) \cite{cer2017semeval}, and Quora Question Pairs (QQP) 
\item Four natural language inference (NLI) tasks
including Multi NLI (MNLI) \cite{williams2018broad}, Question NLI (QNLI) \cite{rajpurkar2016squad}, Recognizing Textual Entailment
(RTE) \cite{dagan2005pascal,giampiccolo2007third}, and Winograd NLI (WNLI) \cite{levesque2012winograd}.
\end{itemize}
\vspace{1mm}

The performance of our FedSplitBERT framework in comparison with baseline methods are evaluated under the non-IID settings, which indicates that data over different clients have distinct label distributions.

\begin{table}[htb]
\centering
\renewcommand{\arraystretch}{1.3}
\caption{The label distributions for each client used to construct non-IID datasets for 3 clients (average score for STSB and percentages of every class for classification tasks).}
\label{tab:noniid_data_distribution}
\begin{tabular}{llll}
\hline
\textbf{Dataset} & \textbf{Client1} & \textbf{Client2} & \textbf{Client3} \\  \hline
\multicolumn{1}{l}{\textbf{MRPC}}   & 95\%/5\%         & 75\%/25\%        & 25\%/75\%        \\
\multicolumn{1}{l}{\textbf{CoLA}}   & 90.5\%/9.5\%     & 83\%/17\%        & 40\%/60\%        \\
\textbf{QQP}       & 70\%/30\%        & 37\%/63\%        & 13\%/87\%        \\
\textbf{MNLI}      & 86\%/5\%/9\%     & 5\%/90\%/5\%     & 5\%/5\%/90\%     \\
\textbf{STSB}      & 0.93             & 2.90             & 4.26     \\
\textbf{Other}     & 80\%/20\%        & 50\%/50\%        & 20\%/80\%  \\\hline      
\end{tabular}
\end{table}

Prior to running experiments, we create synthetic client data from the GLUE benchmark.
Throughout our experiments, every client has approximately the same number of data, \emph{i.e.}, $|\mathcal{D}|/K$, where $|\mathcal{D}|$ is the total sample size and $K$ is the client number.
We manually construct synthetic data for $K$ clients to vary the distribution of data labels across each client. To achieve this goal, we simulate synthetic data for all clients according to a distribution scheme specific to each dataset. Table \ref{tab:noniid_data_distribution} displays the data distribution schemes of 9 datasets for 3 clients. In this experiment, 8 of 9 datasets are classification tasks, except STSB, which is a regression task. Therefore, in Table \ref{tab:noniid_data_distribution} we present average scores for three clients for the STSB dataset. Specifically, for STSB we sorted the samples by their scores, then split the whole dataset to $K$ parts in order, each part has $|\mathcal{D}|/K$ examples. So client 1, 2 and 3 have different distribution of scores, which ranges in 0.0-2.0, 2.0-3.6 and 3.6-5.0, respectively. For other datasets, we randomly sample data without replacement according to the given label probabilities in this table for each client.

For the experiments with 10 clients, we construct synthetic data for each client in a similar manner by varying the percentage of positive labels from approximately 90\% to 0\% with a step-size of 10\%. Table \ref{tab:noniid_data_distribution_10} shows the data construction details of experiment with 10 clients. Due to the task MRPC has relatively small number of data samples and imbalance positive and negative label distribution, its synthetic non-iid dataset cannot strictly follow the step-size of 10\%, we set different distribution for each client.

\begin{table}[htb]
\centering
\renewcommand{\arraystretch}{1.3}
\caption{The label distributions for each client used to construct non-IID datasets for 10 clients on MPRC, SST-2, QQP and QNLI tasks.}
\label{tab:noniid_data_distribution_10}
\begin{tabular}{lcl}
\hline
\textbf{Dataset} & \textbf{QQP}, \textbf{SST-2},\textbf{QNLI} & \textbf{MRPC} \\  \hline
\multicolumn{1}{l}{\textbf{Client1}}   & 90\%/10\%      & 92\%/8\%      \\
\multicolumn{1}{l}{\textbf{Client2}}   & 80\%/20\%      & 86\%/14\%    \\
\textbf{Client3}       & 70\%/30\%       &  71\%/29\%   \\
\textbf{Client4}      & 60\%/40\%        &  66\%/34\%   \\
\textbf{Client5}      & 50\%/50\%        & 50\%/50\%   \\
\textbf{Client6}     & 40\%/60\%.        & 44\%/56\%  \\
\textbf{Client7}     & 30\%/70\%         & 35\%/65\%  \\
\textbf{Client8}     & 20\%/80\%.        & 28\%/72\%  \\
\textbf{Client9}     & 10\%/90\%.        & 12\%/88\%  \\
\textbf{Client10}     & 2\%/98\%         & 51\%/48\%  \\
\hline      
\end{tabular}
\end{table}

\begin{table*}[bht]
\caption{Performance of FedAvg, FedProx, FedAdam and FedSplitBERT on 9 GLUE datasets under non-IID setting over 3 clients.  Last row, we present the performance of FedSplitBERT with quantization (FedSplitBERT+qat). 
}
\label{tab:noniid_result}
\renewcommand{\arraystretch}{1.3}
\resizebox{\linewidth}{!}{
\begin{tabular}{lcllllcllllc}
\hline
\multirow{2}{*}{\textbf{Dataset}} & \textbf{MRPC}   & \textbf{COLA}  & \textbf{SST-2}  & \textbf{QQP}  & \textbf{MNLI-(m/mm)}   & \textbf{RTE}    & \textbf{QNLI}   & \textbf{WNLI}   & \textbf{STSB}  & \multirow{2}{*}{\textbf{Average}} \\ \cline{2-10} 
   & F1/Acc    & Mcc    & Acc    & Acc    & Acc    & F1/Acc    & Acc    & Acc    & Pear. corr &   \\\hline
\multicolumn{1}{l}{FedAvg}  & 76.65/83.96 & 48.81 & 88.65/90.77 & 88.61 & 78.00/76.66    & 57.66/61.68     & 87.02 & 54.43  & 64.36 &  72.25   \\
\multicolumn{1}{l}{FedProx} & 76.67/82.31 & 46.45 & 88.92/91.17 & 88.12 & 79.89/77.14 & 57.03/64.81     & 86.83 & 52.23 & 69.71 &  72.87   \\
\multicolumn{1}{l}{FedAdam} & 76.99/85.41 & 45.29 & 89.50/91.65 & 90.12 & 80.00/80.66    & 60.28/72.14     & 88.36 & 65.29 & 70.72 &  75.61   \\\hline
\multicolumn{1}{l}{FedSplitBERT} & \textbf{79.97/86.15} & \textbf{51.17} & \textbf{90.12/93.78} & \textbf{90.54} & \textbf{82.24/78.86} & \textbf{62.78/73.11} & \textbf{90.65} & \textbf{68.90} & 71.15  &  \textbf{77.94}  \\
\multicolumn{1}{l}{\begin{tabular}[c]{@{}l@{}}FedSplitBERT +qat\end{tabular}} 
  & 77.95/86.27 & 49.23 & 90.29/91.04 & 90.49 & 80.32/78.20 & 59.68/71.89 & 89.41 & 67.49 & \textbf{71.94}  & 76.64 \\ \hline

\end{tabular}
}
\end{table*}

\begin{table*}[htb]
\caption{Performance of FedSplitBERT under non-IID setting over 3 clients. The critical layer $c$ is fixed at 0, 4, 6, 8, 10 and 12 (same as FedAvg).}
\label{tab:diff_c_results}
\renewcommand{\arraystretch}{1.3}
\resizebox{\linewidth}{!}{
\begin{tabular}{lcllllcllllc}
\hline
\multirow{2}{*}{\textbf{Dataset}} & \multirow{2}{*}{\textbf{c}} &\textbf{MRPC}   & \textbf{COLA}  & \textbf{SST-2}  & \textbf{QQP}  & \textbf{MNLI-(m/mm)}   & \textbf{RTE}    & \textbf{QNLI}   & \textbf{WNLI}   & \textbf{STSB} & \multirow{2}{*}{\textbf{Average} }   \\ \cline{3-11} 
 &   & F1/Acc    & Mcc    & Acc    & Acc    & Acc    & Acc    & Acc    & Acc    & Pear. corr &   \\\hline

\multirow{5}{*}{\begin{tabular}[c]{@{}l@{}}FedSplitBERT\end{tabular}} 
  & 12 & 76.65/83.96 & 48.81 & 90.77 & 88.61 & 78.00/76.66    & 61.68 & 87.02 & 54.43  & 64.36 &  72.25   \\
  & 10 & 78.98/86.03 & 49.24 & \textbf{93.78} & 89.19 & \textbf{82.24/78.86} & \textbf{74.11} & \textbf{90.65} & 68.90 & \textbf{71.15}  & 77.58   \\
  & 8 & \textbf{79.97/86.15}  & 50.33 & 93.27 & 90.18 & 80.70/74.71  & 73.11 & 89.72 & \textbf{71.31} & 70.94 &  \textbf{77.72} \\
  & 6 & 79.57/86.85 & \textbf{51.17} & 93.43 & \textbf{90.54} & 80.31/71.83 & 73.76 & 89.67 & 68.46 & 70.35 &  77.47 \\
  & 4 & 77.60/85.96 & 43.68 & 91.22 & 90.35 & 78.42/70.27 & 72.61 & 89.51 & 61.65 & 70.58  &  75.06 \\    
  & 0 & 75.01/86.21 & 43.40 & 89.72 & 90.01 & 77.96/69.28 & 69.37 & 88.49 & 50.01  & 68.30 & 72.47 \\    \hline

\end{tabular}
}
\end{table*}

\subsection{Experimental Settings}

\noindent \textbf{Test Set \& Implementation}
Since we need the true labels to construct synthetic client data,
we merge the training and development sets of GLUE benchmarks as the whole datasets, and on each client we split the dataset by 80\% and 20\% for training and test sets. When performing FL experiments, training data at all clients are used to train the model and our FedSplitBERT methods are evaluated on the local test sets of clients. We cannot utilize the original test sets in GLUE data as it is impossible to construct heterogeneous test data consistent to the training data without label.

For each dataset, we employ five different federated methods: FedAvg, FedProx, FedAdam, and FedSplitBERT with/without quantization, and compare their results. We used the pre-trained model 'bert-base-uncased' (Layers=12, Hidden size=768, Self-Attention Heads=12) in our research. Then, we quantized 32-bit weights to 16-bit and reverted back with basic Pytorch Tensor data structure operations \cite{paszke2019pytorch}. All our experiments were conducted on NVIDIA Tesla V100.

\vspace{3.5pt}
\noindent \textbf{Hyperparameters \& Evaluation Metrics}
For each task, the local learning rate $\eta$ takes a value in the set \{2e-5, 3e-5, 4e-5, 5e-5\}, and the local fine-tuning epochs ranging from 3 to 9, the batch size of local training is chosen from \{8, 16, 32\}, with the input padding length fixed at 128. We used the hyper-parameters of FedAdam from \cite{reddi2021adaptive}. We conducted the experiments under 3 and 10 clients federated setting and tuned the hyper-parameter $\lambda$ for FedProx, critical layers $c$ for FedSplitBERT. At each iteration, all clients are participated for model aggregation in our experiments.

We report the average Matthew's corr for CoLA, average Pearson for STSB, both average F1 score and average Accuracy for RTE and MRPC, and average Accuracy for the rest of datasets. For FedSplitBERT, we evaluate the accuracy under the local test setting, following the method in \cite{liang2020think}.

\subsection{Results and Analysis}


\paragraph{Heterogeneity}
Table \ref{tab:noniid_result} and \ref{tab:noniid_result_10c} depict the performance of various FL methods under non-IID settings over 3 and 10 clients, respectively. In 10-client setting, we test the baseline methods (FedAvg, FedAdam, and FedProx) and FedSplitBERT on 4 datasets as it is very resource demanding for training up to 10 BERT models at the same time. 
Here, the baseline methods train one shared BERT model for all clients. By contrast, we implement FedSplitBERT by setting the critical layer $c$ at 5 values: 0, 4, 6, 8, 10. When $c=0$, it means that there is no shared global model among clients, and each client trains its own BERT model with its local isolated data. Due to the overfitting on local data and learning from less data than global model, FedSplitBERT ($c=0$) performs the worst among FedSplitBERT methods, and some times even worse than FedAvg. 

\begin{table}[bht]
\caption{Performance of FedAvg, FedProx, FedAdam, and FedSplitBERT on 4 datasets under non-IID setting over 10 clients. The communication rounds $t$ for MPRC is set to 6, $t=9$ for SST-2 and $t=3$ for QQP and QNLI.
}
\label{tab:noniid_result_10c}
\renewcommand{\arraystretch}{1.3}
\resizebox{\linewidth}{!}{
\begin{tabular}{lllllc}
\hline
\multirow{2}{*}{\textbf{Dataset}}  &\textbf{MRPC}   & \textbf{SST-2}  & \textbf{QQP}  & \textbf{QNLI}  & \multirow{2}{*}{\textbf{Average} }   \\ \cline{2-5} 
& Acc    & Acc    & Acc      & Acc     \\\hline
\multicolumn{1}{l}{FedAvg} & 62.95 & 92.36 & 84.56 & 82.84  & 80.68   \\
\multicolumn{1}{l}{FedProx}& 64.15 & 92.25 & 86.03 & 81.16 & 80.90    \\
\multicolumn{1}{l}{FedAdam}& 67.20 & 93.15 & 86.32 & 84.22  & 82.72   \\\hline
\multicolumn{1}{l}{FedSplitBERT} & 79.86 & \textbf{94.92} & \textbf{91.35} & 89.14  &  \textbf{88.81}  \\
\multicolumn{1}{l}{\begin{tabular}[c]{@{}l@{}}FedSplitBERT +qat\end{tabular}} & \textbf{79.96} & 94.13 & 90.76 & \textbf{89.24} & 88.52 \\ \hline
\end{tabular}}
\end{table}

\begin{figure}[htb]
    \centering
    \includegraphics[width=0.5\textwidth]{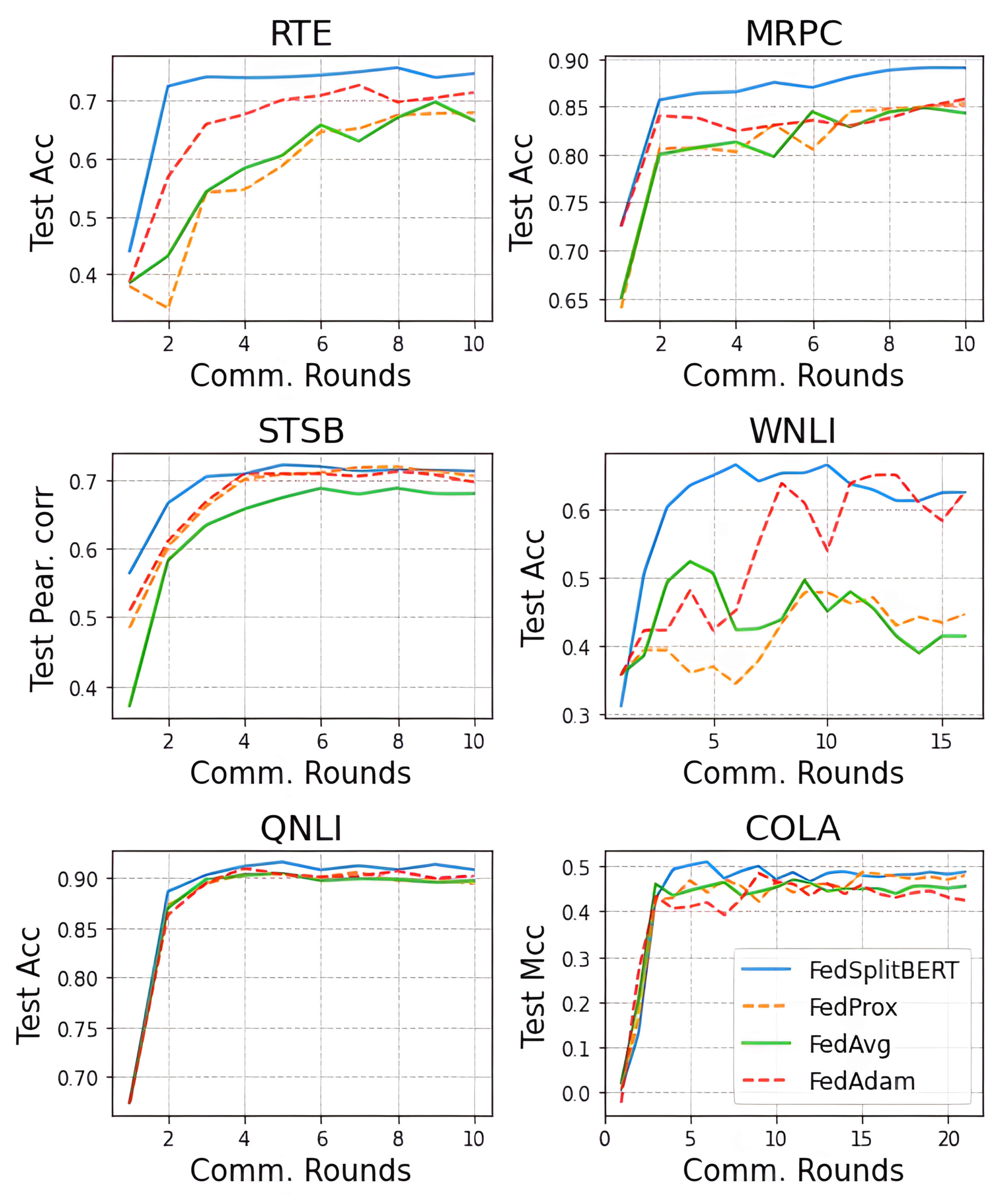}
    \caption{The test accuracy versus communication rounds of the baseline methods and FedSplitBERT without quantization on 6 datasets.}
    \label{fig:acc_v_rounds}
\end{figure}

\begin{figure}[htb]
    \centering
    \includegraphics[width=0.5\textwidth]{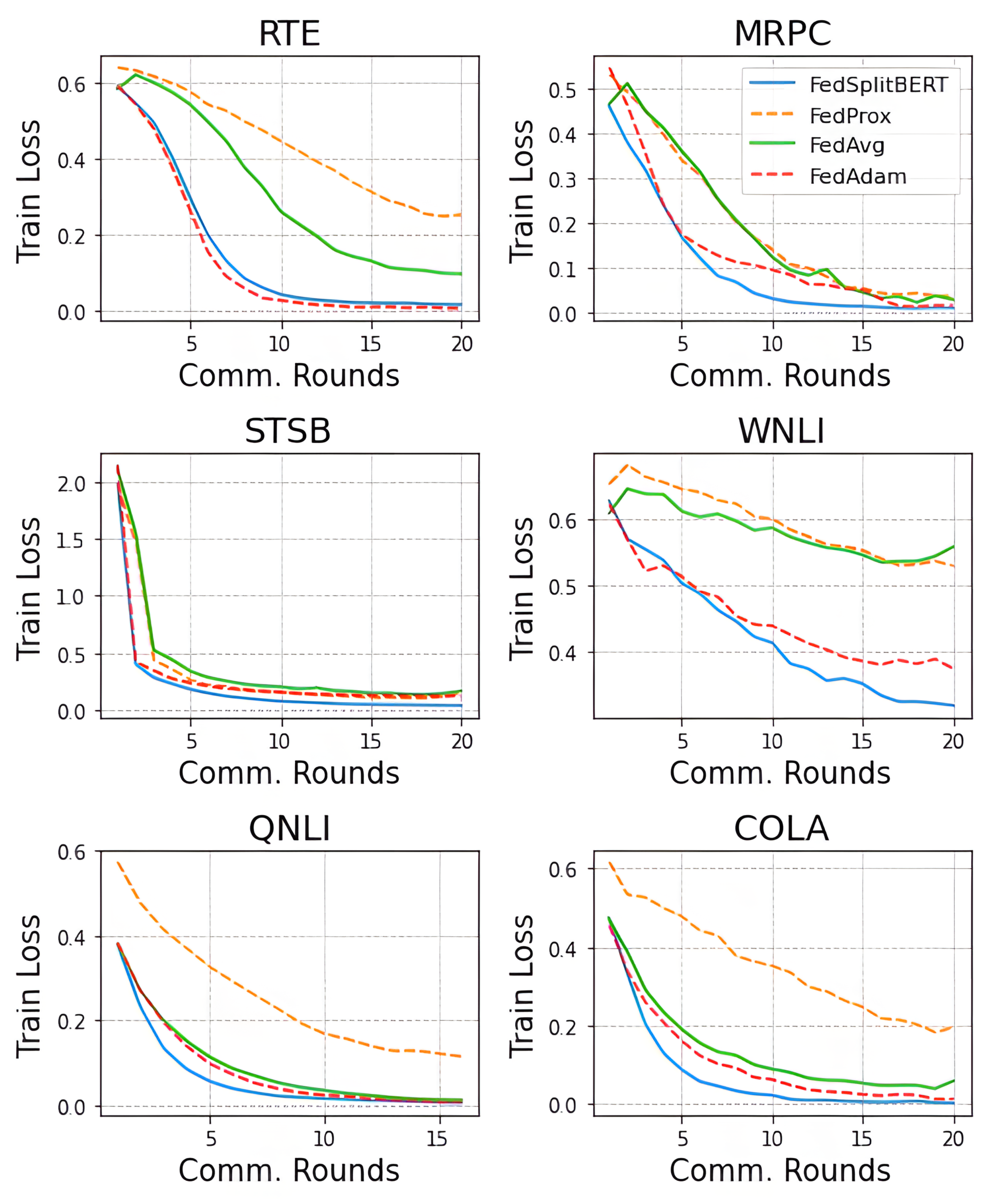}
    \caption{The training loss versus communication rounds of the baseline methods and FedSplitBERT without quantization on 6 datasets.}
    \label{fig:loss_v_rounds}
\end{figure}

\begin{figure}[htb]
    \centering
    \includegraphics[width=0.5\textwidth]{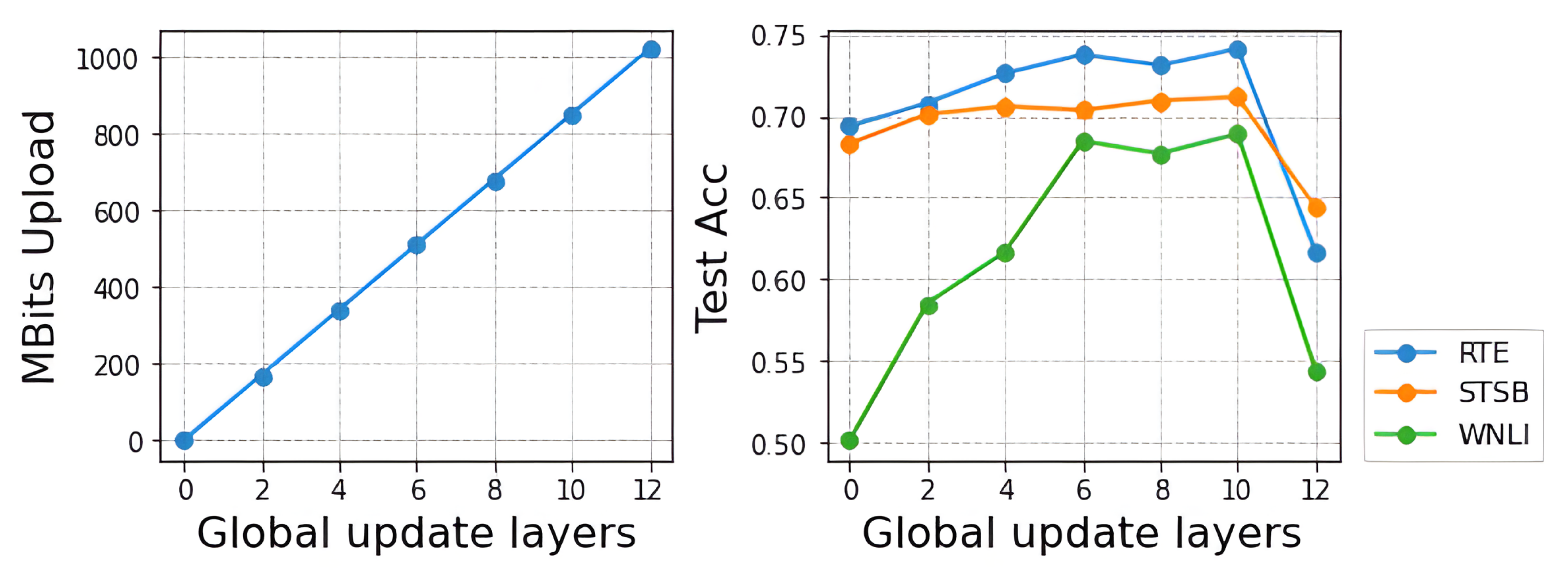}
    \caption{Left plot reports the communication cost of each round in MBits versus different $c$ of FedSplitBERT when client number $K$=3. Right plot displays test accuracy versus the number of global layers $c$ on 3 datasets.}
    \label{fig:layers_acc_loss}
\end{figure}

Our FedSplitBERT and FedSplitBERT+qat ($c=6$) outperform baseline methods (FedAvg, FedProx and FedAdam) significantly on all datasets, by an average score improvement of nearly 4\% and at least 1\% on single task. 
Also, Table \ref{tab:diff_c_results} reports the performance of FedSplitBERT under different choices of critical layer c (6 values).The study \cite{jawahar2019does} shows that BERT captures phrase-level information in the lower layers, syntactic information in the middle layers, semantic information at the top layers. Our experiments show that when layer $c$ ranges from 6 to 8, FedSplitBERT has better performance than $c<6$ and $c=12$ on the linguistic related classification task (CoLA) and sentence similarity tasks, such as MRPC, STSB and QQP. For nature language inference tasks, MNLI, QNLI, WNLI and RTE, a higher c yields better performance. These results imply a finding that our global encoders for BERT model can learn the general features, which is important for these tasks, while deeper layers corresponding to the local model in FedSplitBERT learn high-level feature representations and semantic information. The layer c can be be selected according to the type of task. Our FedSplitBERT model with $c=6, 8, 10$ can both utilize more data to learn general features and keep the ability to capture local characteristics (Fig. \ref{fig:layers_acc_loss}). Therefore, a general suggestion for 
selecting the critical layer c is to try 6, 8 and 10. The value of c balances the model capacity and communication cost, so different datasets require different values of c.

FedAvg, FedProx and FedAdam have critical layer $c=12$, so they fail to capture data heterogeneity. Under this setting, one global BERT model cannot depict the data features and heterogeneous distribution of different clients. On the other hand, when $c=0$, the FedSplitBERT is equivalent to isolate training, with all the layers locally updated. In this case, the model only learns from client's local data, leading to overfitting. Our FedSplitBERT with appropriate choice of critical layer addresses these two problems and achieves better performance as well as faster convergence. 

In addition, we report the fine-tuning results of our approach over 16-bit quantization (FedSplitBERT+qat). It achieves a similar performance with FedSplitBERT ($c=6$) with a very low communication cost. FedSplitBERT+qat performs the best on MRPC and QNLI datasets under the 10 clients setting and STSB under the 3 clients setting, which may be due to the decreased over-fitting by truncating the weights.

\paragraph{Communication Complexity}
Communication cost is an important metric for federated
learning. Two factors concur in reducing communication complexity: the rate of convergence at a target accuracy, which determines the number of rounds, and the amount of model parameters transferred at each round.
We show in Fig. \ref{fig:acc_v_rounds} and \ref{fig:loss_v_rounds} the convergence analysis of the training loss and test accuracy over the number of rounds communications for the non-IID setting across 6 datasets: RTE, MRPC, STSB, WNLI, QNLI and COLA. Our methods converge more quickly than the FedAvg and FedProx, achieving a higher accuracy and lower training loss across all datasets.

\begin{figure}[htb]
    \centering
    \includegraphics[width=0.49\textwidth]{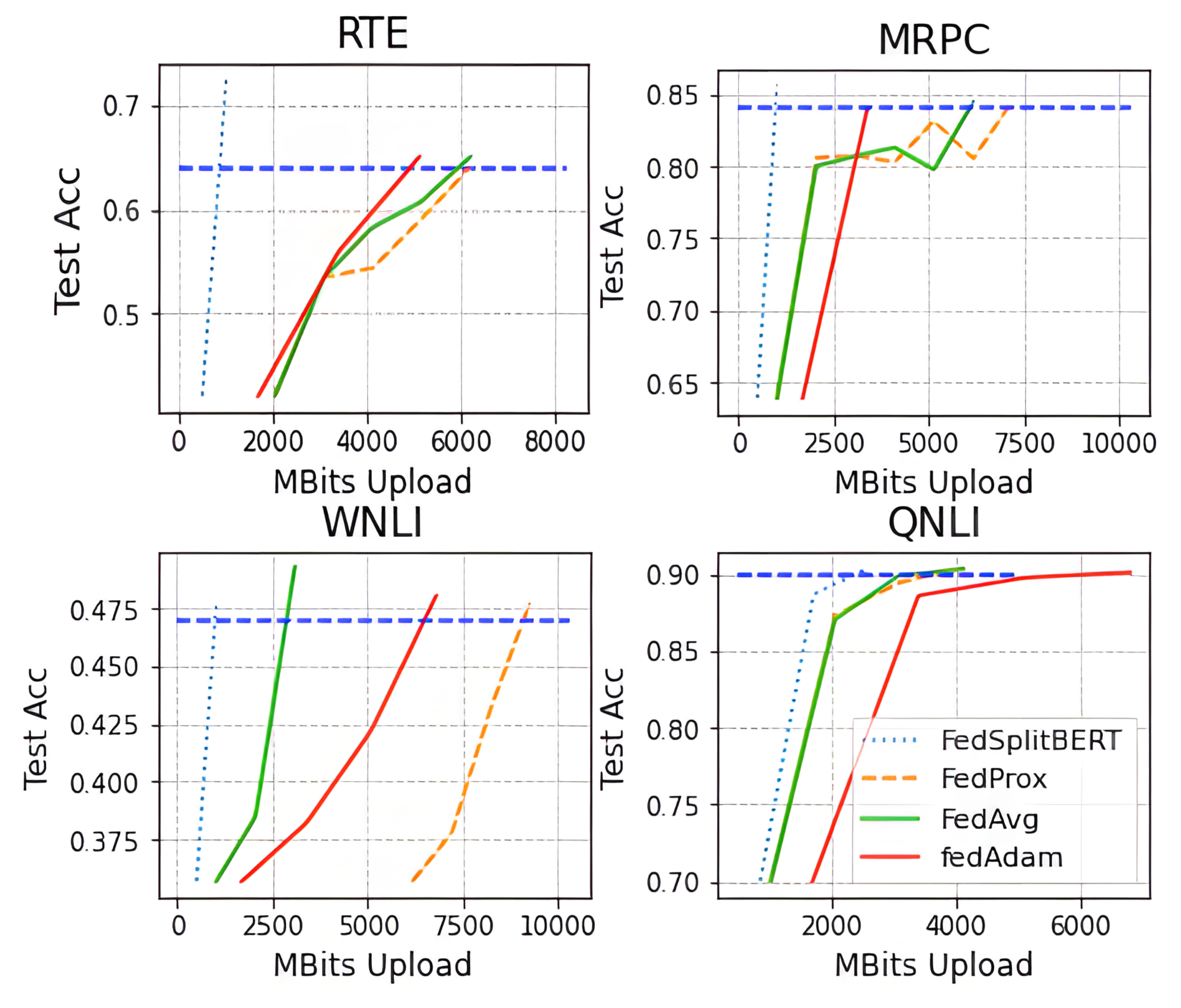}
    \caption{Communication cost in MBits when baseline methods and FedSplitBERT attain a particular accuracy on four datasets(non-iid): RTE(64\%), MRPC(84\%), WNLI(47\%) and QNLI(90\%). Client number $K$=3.}
    \label{fig:Mbuploads}
\end{figure}

\begin{table}[htb]
\caption{First column shows uploading and downloading parameter size in Gigabyte in each communication round $t$. Last two columns demonstrate the communication cost and gain for baseline methods and FedSplitBERT with quantization(+qat) to achieve a test accuracy of 84\% on MRPC. Note client number $K=3$, critical layer $c=6$.}
\label{tab:comm.size}
\renewcommand{\arraystretch}{1.2}
\resizebox{\columnwidth}{!}{
\begin{tabular}{lccccc}
\hline
\textbf{Method} &
{\begin{tabular}[c]{@{}c@{}}\textbf{Para. size}\\ \textbf{(GB)}\end{tabular}} &
{\begin{tabular}[c]{@{}c@{}}\textbf{Comm. Cost}\\ \textbf{(GB)}\end{tabular}} & 
\textbf{Comm. gain} \\ \hline
FedAvg/FedProx   & 1.03 & 6.18 & 1$\times$   \\
FedAdam          & 1.7  & 3.4  & 1.8$\times$ \\
FedSplitBERT     & 0.51 & 1.02 & 6.1$\times$ \\
FedSplitBERT+qat & 0.26 & 0.52 & 11.9$\times$\\\hline
\end{tabular}}
\end{table}

We demonstrate the necessary size of uploading and downloading parameters of the baseline methods and our proposed framework at each round over 3 clients and critical layer $c=6$. The communication cost to attain the same performance(84\% accuracy) for these methods on MRPC dataset are reported in Table \ref{tab:comm.size}. The minimum transferred parameter size of FedSplitBERT is 0.26 GB with quantization, 4 times lower than FedAvg and FedProx, 6 times lower than FedAdam. Furthermore, since FedSplitBERT has a faster convergence rate than fully generic model on heterogeneous data, the communication complexity decreases along with the number of parameter-transferring rounds. 

Fig.\ref{fig:Mbuploads} displays the communication cost in terms of uploading size in MBits when all methods achieving a particular level of test accuracy on four datasets. This figure shows the trend of faster convergence of FedSplitBERT than FedAvg, Fedprox and FedAdam on 4 different datasets. On RTE tasks, the dotted blue line shows that our FedSplitBERT without quantization converges within only 3 communication rounds which costs nearly 1000 MBits to exceed the 64\% accuracy, while baseline methods need at least 5000 MBits communication overhead. On MRPC and WNLI tasks, FedSplitBERT also attains the performance threshold with less than 1GB communication cost. However, baseline methods need at least 2$\times$ communication cost to achieve the same performance. With quantization the global layers, our methods obtain 11.9$\times$ communication gain, when achieve the comparable performance to FedAvg (non-IID).

\section{Conclusion and Discussion}

In this paper, we propose a framework, FedSplitBERT, to address the heterogeneous data in federated BERT fine-tuning. We evaluate our methods on the GLUE benchmarks and we find that our methods exceed FedAvg, FedProx and FedAdam by a significant margin for heterogeneous clients. Through ablation studies, we find that the critical layer $c$ for FedSplitBERT can be tuned according to the type of the downstream task.
Due to the huge size of BERT model, we also investigate quantization to further reduce the communication cost, which could reduce the communication cost by $11.9\times$. Our method is easy-to-implement, and very effective to heterogeneous data. Our framework is also compatible to many existing and incoming FL aggregation algorithms, like FedYogi, etc.
In the future, we can study more NLP tasks with BERT in federated setting, like named entity recognition, etc. Another important problem is how to reduce the communication cost due to the increasing size of NLP models. We have studied the quantization method in this paper, and many other efficient weight compression methods can be employed.

\section*{Acknowledgment}
This paper is supported by the Key Research and Development Program of Guangdong Province under grant No.2021B0101400003. Corresponding author is Jianzong Wang from Ping An Technology (Shenzhen) Co., Ltd (wangjianzong347@pingan.com.cn).
\bibliographystyle{IEEEtran}
\bibliography{IEEEfedbert}

\begin{thebibliography}{10}
\providecommand{\url}[1]{#1}
\csname url@samestyle\endcsname
\providecommand{\newblock}{\relax}
\providecommand{\bibinfo}[2]{#2}
\providecommand{\BIBentrySTDinterwordspacing}{\spaceskip=0pt\relax}
\providecommand{\BIBentryALTinterwordstretchfactor}{4}
\providecommand{\BIBentryALTinterwordspacing}{\spaceskip=\fontdimen2\font plus
\BIBentryALTinterwordstretchfactor\fontdimen3\font minus
  \fontdimen4\font\relax}
\providecommand{\BIBforeignlanguage}[2]{{%
\expandafter\ifx\csname l@#1\endcsname\relax
\typeout{** WARNING: IEEEtran.bst: No hyphenation pattern has been}%
\typeout{** loaded for the language `#1'. Using the pattern for}%
\typeout{** the default language instead.}%
\else
\language=\csname l@#1\endcsname
\fi
#2}}
\providecommand{\BIBdecl}{\relax}
\BIBdecl

\bibitem{peters2018deep}
M.~E. Peters, M.~Neumann, M.~Iyyer, M.~Gardner, C.~Clark, K.~Lee, and
  L.~Zettlemoyer, ``Deep contextualized word representations,'' in
  \emph{Proceedings of the 2018 Conference of the North {A}merican Chapter of
  the Association for Computational Linguistics: Human Language Technologies,
  Volume 1 (Long Papers)}, Jun. 2018, pp. 2227--2237.

\bibitem{brown2020language}
T.~Brown, B.~Mann, N.~Ryder, M.~Subbiah, J.~D. Kaplan, P.~Dhariwal,
  A.~Neelakantan, P.~Shyam, and e.~a. Sastry, Girish, ``Language models are
  few-shot learners,'' in \emph{Advances in Neural Information Processing
  Systems}, H.~Larochelle, M.~Ranzato, R.~Hadsell, M.~F. Balcan, and H.~Lin,
  Eds., vol.~33.\hskip 1em plus 0.5em minus 0.4em\relax Curran Associates,
  Inc., 2020, pp. 1877--1901.

\bibitem{devlin2019bert}
J.~Devlin, M.-W. Chang, K.~Lee, and K.~Toutanova, ``Bert: Pre-training of deep
  bidirectional transformers for language understanding,'' in \emph{Proceedings
  of the 2019 Conference of the North American Chapter of the Association for
  Computational Linguistics: Human Language Technologies, Volume 1 (Long and
  Short Papers)}, 2019, pp. 4171--4186.

\bibitem{wang2016attention}
Y.~Wang, M.~Huang, X.~Zhu, and L.~Zhao, ``Attention-based lstm for aspect-level
  sentiment classification,'' in \emph{Proceedings of the 2016 conference on
  empirical methods in natural language processing}, 2016, pp. 606--615.

\bibitem{williams2018broad}
A.~Williams, N.~Nangia, and S.~Bowman, ``A broad-coverage challenge corpus for
  sentence understanding through inference,'' in \emph{Proceedings of the 2018
  Conference of the North American Chapter of the Association for Computational
  Linguistics: Human Language Technologies, Volume 1 (Long Papers)}, 2018, pp.
  1112--1122.

\bibitem{lai2017race}
G.~Lai, Q.~Xie, H.~Liu, Y.~Yang, and E.~Hovy, ``{RACE}: Large-scale reading
  comprehension dataset from examinations,'' in \emph{Proceedings of the 2017
  Conference on Empirical Methods in Natural Language Processing}.\hskip 1em
  plus 0.5em minus 0.4em\relax Copenhagen, Denmark: Association for
  Computational Linguistics, Sep. 2017, pp. 785--794.

\bibitem{xu2021federated}
J.~Xu, B.~S. Glicksberg, C.~Su, P.~Walker, J.~Bian, and F.~Wang, ``Federated
  learning for healthcare informatics,'' \emph{Journal of Healthcare
  Informatics Research}, vol.~5, no.~1, pp. 1--19, 2021.

\bibitem{voigt2017eu}
P.~Voigt and A.~Von~dem Bussche, ``The eu general data protection regulation
  (gdpr),'' \emph{A Practical Guide, 1st Ed., Cham: Springer International
  Publishing}, vol.~10, p. 3152676, 2017.

\bibitem{mcmahan2017communication}
B.~McMahan, E.~Moore, D.~Ramage, S.~Hampson, and B.~A. y~Arcas,
  ``Communication-efficient learning of deep networks from decentralized
  data,'' in \emph{Artificial Intelligence and Statistics}.\hskip 1em plus
  0.5em minus 0.4em\relax PMLR, 2017, pp. 1273--1282.

\bibitem{kairouz2021advances}
P.~Kairouz, H.~B. McMahan, B.~Avent, A.~Bellet, M.~Bennis, A.~N. Bhagoji,
  K.~Bonawitz, Z.~Charles, G.~Cormode, R.~Cummings \emph{et~al.}, ``Advances
  and open problems in federated learning,'' \emph{Foundations and
  Trends{\textregistered} in Machine Learning}, vol.~14, no. 1--2, pp. 1--210,
  2021.

\bibitem{huang2020texthide}
Y.~Huang, Z.~Song, D.~Chen, K.~Li, and S.~Arora, ``{T}ext{H}ide: Tackling data
  privacy in language understanding tasks,'' in \emph{Findings of the
  Association for Computational Linguistics: EMNLP 2020}, 2020, pp. 1368--1382.

\bibitem{liu2020federated}
D.~Liu and T.~Miller, ``Federated pretraining and fine tuning of bert using
  clinical notes from multiple silos,'' \emph{AI for Public Health Workshop at
  ICLR’21}, 2021.

\bibitem{hilmkil2021scaling}
A.~Hilmkil, S.~Callh, M.~Barbieri, L.~R. S{\"u}tfeld, E.~L. Zec, and O.~Mogren,
  ``Scaling federated learning for fine-tuning of large language models,'' in
  \emph{International Conference on Applications of Natural Language to
  Information Systems}.\hskip 1em plus 0.5em minus 0.4em\relax Springer, 2021,
  pp. 15--23.

\bibitem{wang2021modeling}
J.~Wang, Z.~Huang, L.~Kong, D.~Li, and J.~Xiao, ``Modeling without sharing
  privacy: Federated neural machine translation,'' in \emph{International
  Conference on Web Information Systems Engineering}.\hskip 1em plus 0.5em
  minus 0.4em\relax Springer, 2021, pp. 216--223.

\bibitem{yang2020heterogeneity}
C.~Yang, Q.~Wang, M.~Xu, Z.~Chen, Y.~L. Kaigui~Bian, and X.~Liu,
  ``Characterizing impacts of heterogeneity in federated learning upon
  large-scale smartphone data.''\hskip 1em plus 0.5em minus 0.4em\relax
  Ljubljana, Slovenia.ACM, New York, NY, USA, 12 pages.: the Web Conference
  2021 (WWW ’21), April 19–23, 2021.

\bibitem{li2020federated}
T.~Li, A.~K. Sahu, M.~Zaheer, M.~Sanjabi, A.~Talwalkar, and V.~Smith,
  ``Federated optimization in heterogeneous networks,'' \emph{Proceedings of
  Machine Learning and Systems}, vol.~2, pp. 429--450, 2020.

\bibitem{reddi2021adaptive}
S.~J. Reddi, Z.~Charles, M.~Zaheer, Z.~Garrett, K.~Rush, J.~Kone{\v{c}}n{\'y},
  S.~Kumar, and H.~B. McMahan, ``Adaptive federated optimization,'' in
  \emph{International Conference on Learning Representations}, 2021.

\bibitem{acar2021federated}
D.~A.~E. Acar, Y.~Zhao, R.~Matas, M.~Mattina, P.~Whatmough, and V.~Saligrama,
  ``Federated learning based on dynamic regularization,'' in
  \emph{International Conference on Learning Representations}, 2021.

\bibitem{FedSmart111}
A.~He, J.~Wang, Z.~Huang, and J.~Xiao, ``Fedsmart: An auto updating federated
  learning optimization mechanism,'' in \emph{Web and Big Data: 4th
  International Joint Conference, APWeb-WAIM 2020, Tianjin, China, Proceedings,
  Part I}, Berlin, Heidelberg, 2020, p. 716–724.

\bibitem{liang2020think}
P.~P. Liang, T.~Liu, L.~Ziyin, R.~Salakhutdinov, and L.-P. Morency, ``Think
  locally, act globally: Federated learning with local and global
  representations,'' \emph{NeurIPS 2019 Workshop on Federated Learning
  distinguished student paper award}, 2019.

\bibitem{yang2020heterogeneous}
L.~Yang, C.~Beliard, and D.~Rossi, ``Heterogeneous data-aware federated
  learning,'' \emph{IJCAI 2020 Federated Learning Workshop}, 2020.

\bibitem{collins2021exploiting}
L.~Collins, H.~Hassani, A.~Mokhtari, and S.~Shakkottai, ``Exploiting shared
  representations for personalized federated learning,'' in \emph{International
  Conference on Machine Learning}.\hskip 1em plus 0.5em minus 0.4em\relax PMLR,
  2021, pp. 2089--2099.

\bibitem{dinh2020personalized}
C.~T.~Dinh, N.~Tran, and J.~Nguyen, ``Personalized federated learning with
  moreau envelopes,'' in \emph{Advances in Neural Information Processing
  Systems}, H.~Larochelle, M.~Ranzato, R.~Hadsell, M.~F. Balcan, and H.~Lin,
  Eds., vol.~33.\hskip 1em plus 0.5em minus 0.4em\relax Curran Associates,
  Inc., 2020, pp. 21\,394--21\,405.

\bibitem{si2020students}
\BIBentryALTinterwordspacing
S.~Si, R.~Wang, J.~Wosik, H.~Zhang, D.~Dov, G.~Wang, and L.~Carin, ``Students
  need more attention: Bert-based attention model for small data with
  application to automatic patient message triage,'' in \emph{Proceedings of
  the 5th Machine Learning for Healthcare Conference}, ser. Proceedings of
  Machine Learning Research, F.~Doshi-Velez, J.~Fackler, K.~Jung, D.~Kale,
  R.~Ranganath, B.~Wallace, and J.~Wiens, Eds., vol. 126.\hskip 1em plus 0.5em
  minus 0.4em\relax PMLR, 07--08 Aug 2020, pp. 436--456. [Online]. Available:
  \url{https://proceedings.mlr.press/v126/si20a.html}
\BIBentrySTDinterwordspacing

\bibitem{wang2020integrating}
\BIBentryALTinterwordspacing
R.~Wang, S.~Si, G.~Wang, L.~Zhang, L.~Carin, and R.~Henao, ``Integrating task
  specific information into pretrained language models for low resource fine
  tuning,'' in \emph{Findings of the Association for Computational Linguistics:
  EMNLP 2020}.\hskip 1em plus 0.5em minus 0.4em\relax Online: Association for
  Computational Linguistics, Nov. 2020, pp. 3181--3186. [Online]. Available:
  \url{https://aclanthology.org/2020.findings-emnlp.285}
\BIBentrySTDinterwordspacing

\bibitem{shen2020q}
S.~Shen, Z.~Dong, J.~Ye, L.~Ma, Z.~Yao, A.~Gholami, M.~W. Mahoney, and
  K.~Keutzer, ``Q-bert: Hessian based ultra low precision quantization of
  bert,'' in \emph{Proceedings of the AAAI Conference on Artificial
  Intelligence}, vol.~34, no.~05, 2020, pp. 8815--8821.

\bibitem{liu2021hardware}
Z.~Liu, G.~Li, and J.~Cheng, ``Hardware acceleration of fully quantized bert
  for efficient natural language processing,'' \emph{Design, Automation \& Test
  in Europe (DATE)}, 2021.

\bibitem{kim2021bert}
S.~Kim, A.~Gholami, Z.~Yao, M.~W. Mahoney, and K.~Keutzer, ``I-bert:
  Integer-only bert quantization,'' \emph{International Conference on Machine
  Learning (Accepted)}, 2021.

\bibitem{zafrir2019q8bert}
O.~Zafrir, G.~Boudoukh, P.~Izsak, and M.~Wasserblat, ``Q8bert: Quantized 8bit
  bert,'' in \emph{2019 Fifth Workshop on Energy Efficient Machine Learning and
  Cognitive Computing-S Edition (EMC2-NIPS)}.\hskip 1em plus 0.5em minus
  0.4em\relax IEEE, 2019, pp. 36--39.

\bibitem{jin2021kdlsq}
J.~Jin, C.~Liang, T.~Wu, L.~Zou, and Z.~Gan, ``Kdlsq-bert: A quantized bert
  combining knowledge distillation with learned step size quantization,''
  \emph{arXiv preprint arXiv:2101.05938}, 2021.

\bibitem{reisizadeh2020fedpaq}
A.~Reisizadeh, A.~Mokhtari, H.~Hassani, A.~Jadbabaie, and R.~Pedarsani,
  ``Fedpaq: A communication-efficient federated learning method with periodic
  averaging and quantization,'' in \emph{International Conference on Artificial
  Intelligence and Statistics}.\hskip 1em plus 0.5em minus 0.4em\relax PMLR,
  2020, pp. 2021--2031.

\bibitem{jawahar2019does}
G.~Jawahar, B.~Sagot, and D.~Seddah, ``What does bert learn about the structure
  of language?'' in \emph{ACL 2019-57th Annual Meeting of the Association for
  Computational Linguistics}, 2019.

\bibitem{wang2018glue}
A.~Wang, A.~Singh, J.~Michael, F.~Hill, O.~Levy, and S.~R. Bowman, ``{GLUE}: A
  multi-task benchmark and analysis platform for natural language
  understanding,'' in \emph{International Conference on Learning
  Representations}, 2019.

\bibitem{warstadt2019neural}
A.~Warstadt, A.~Singh, and S.~R. Bowman, ``Neural network acceptability
  judgments,'' \emph{Transactions of the Association for Computational
  Linguistics}, vol.~7, pp. 625--641, Mar. 2019.

\bibitem{socher2013recursive}
R.~Socher, A.~Perelygin, J.~Wu, J.~Chuang, C.~D. Manning, A.~Y. Ng, and
  C.~Potts, ``Recursive deep models for semantic compositionality over a
  sentiment treebank,'' in \emph{Proceedings of the 2013 conference on
  empirical methods in natural language processing}, 2013, pp. 1631--1642.

\bibitem{dolan2005automatically}
W.~B. Dolan and C.~Brockett, ``Automatically constructing a corpus of
  sentential paraphrases,'' in \emph{Proceedings of the Third International
  Workshop on Paraphrasing (IWP2005)}, 2005.

\bibitem{cer2017semeval}
D.~Cer, M.~Diab, E.~Agirre, I.~Lopez-Gazpio, and L.~Specia, ``Semeval-2017 task
  1: Semantic textual similarity-multilingual and cross-lingual focused
  evaluation,'' 2017.

\bibitem{rajpurkar2016squad}
P.~Rajpurkar, J.~Zhang, K.~Lopyrev, and P.~Liang, ``Squad: 100,000+ questions
  for machine comprehension of text,'' in \emph{Proceedings of the 2016
  Conference on Empirical Methods in Natural Language Processing}.\hskip 1em
  plus 0.5em minus 0.4em\relax Austin, Texas: Association for Computational
  Linguistics, 2016, pp. 2383--2392.

\bibitem{dagan2005pascal}
I.~Dagan, O.~Glickman, and B.~Magnini, ``The pascal recognising textual
  entailment challenge,'' in \emph{achine Learning Challenges. Evaluating
  Predictive Uncertainty, Visual Object Classification, and Recognising Tectual
  Entailment}, Springer.\hskip 1em plus 0.5em minus 0.4em\relax Springer Berlin
  Heidelberg, 2006, pp. 177--190.

\bibitem{giampiccolo2007third}
D.~Giampiccolo, B.~Magnini, I.~Dagan, and W.~B. Dolan, ``The third pascal
  recognizing textual entailment challenge,'' in \emph{Proceedings of the
  ACL-PASCAL workshop on textual entailment and paraphrasing}, 2007, pp. 1--9.

\bibitem{levesque2012winograd}
H.~Levesque, E.~Davis, and L.~Morgenstern, ``The winograd schema challenge,''
  in \emph{Thirteenth International Conference on the Principles of Knowledge
  Representation and Reasoning}.\hskip 1em plus 0.5em minus 0.4em\relax
  Citeseer, 2012.

\bibitem{paszke2019pytorch}
A.~Paszke, S.~Gross, F.~Massa, A.~Lerer, J.~Bradbury, G.~Chanan, T.~Killeen,
  Z.~Lin, N.~Gimelshein, L.~Antiga \emph{et~al.}, ``Pytorch: An imperative
  style, high-performance deep learning library,'' in \emph{Advances in Neural
  Information Processing Systems}, 2019, pp. 8024--8035.

\end{thebibliography}


\end{document}